%% file: main.tex
\documentclass[sigconf,10pt]{acmart}

\renewcommand\footnotetextcopyrightpermission[1]{} 
\usepackage{multirow}
\usepackage{balance}
\usepackage{adjustbox}

\usepackage{listings}
\usepackage{balance}

\AtBeginDocument{%
  \providecommand\BibTeX{{%
    \normalfont B\kern-0.5em{\scshape i\kern-0.25em b}\kern-0.8em\TeX}}}

\acmDOI{10.475/123_4}

\acmISBN{123-4567-24-567/08/06}

\setcopyright{acmcopyright}
\copyrightyear{2020}
\acmYear{2020}

\begin{document}

\title{EFloat: Entropy-coded Floating Point Format for Compressing
  Vector Embedding Models}

\author{Rajesh Bordawekar}
\email{bordaw@us.ibm.com}
\affiliation{%
  \institution{IBM T.J. Watson Research Center}
  \city{Yorktown Heights}
  \state{NY}
  \country{USA}}

\author{Bulent Abali}
\email{abali@us.ibm.com}

\affiliation{%
  \institution{IBM T.J. Watson Research Center}
  \city{Yorktown Heights}
  \state{NY}
  \country{USA}}

\author{Ming-Hung Chen}
\email{minghungchen@ibm.com}

\affiliation{%
  \institution{IBM T.J. Watson Research Center}
  \city{Yorktown Heights}
  \state{NY}
  \country{USA}}

\input{abs}

\maketitle
\pagestyle{plain}

\input{intro}
\input{design}
\input{efloat}
\input{impl}
\input{eval}
\input{related}

\input{concl}
\input{ack}

\balance

\bibliographystyle{ACM-Reference-Format}
\bibliography{refs}
\end{document}

%% file: abs.tex
\begin{abstract}
In a large class of deep learning models, including vector embedding
models such as word and database embeddings, we observe that floating point exponent values cluster
around a few unique values, permitting entropy based data
compression. Entropy coding compresses fixed-length values with
variable-length codes, encoding most probable values with fewer bits.
We propose the EFloat compressed floating point number format
that uses a variable field boundary between the exponent
and significand fields.  EFloat uses entropy coding on
exponent values and signs to minimize the average
width of the exponent and sign fields, while preserving the original FP32
exponent range unchanged. Saved bits become part of the significand
field increasing the EFloat numeric precision by 4.3 bits on average
compared to other reduced-precision floating point formats.
EFloat makes 8-bit and even smaller floats
practical without sacrificing the exponent range of a 32-bit
floating point representation.  We currently use the EFloat
format for saving memory capacity and bandwidth consumption of large
vector embedding models such as those used for database embeddings.
Using the RMS error as metric, we
demonstrate that EFloat provides higher accuracy than other floating
point formats with equal bit budget.
The EF12 format with 12-bit budget has less end-to-end application
error than the 16-bit BFloat16.  EF16 with 16-bit budget has an
RMS-error 17 to 35 times less than BF16 RMS-error for a diverse set of embedding models.
When making similarity and dissimilarity queries, using the
NDCG ranking metric, EFloat matches the result quality of prior
floating point representations with larger bit budgets.
\end{abstract}

%% file: intro.tex
\section{Introduction}


As natural language processing (NLP) models expand their capabilities and complexity,
their sizes have been increasing dramatically demanding higher memory capacity and bandwidth.
For example, state-of-the-art transformer-based NLP models such as
BERT~(\citet{vaswani2017attention}), Megatron-LM~(\citet{shoeybi2020megatronlm}), Open AI GPT-3~(\citet{openai-gpt3}),
or Google Switch-C Transformers~(\citet{fedus2021switch}), contain from hundreds
of millions, to even trillion
parameters~(\citet{hoefler:twit,fedus2021switch}).
Although NLP model transformation is
a very active area of research (Section~\ref{sec:related}), its current focus is on model inference scenarios
in which reduced precision and integer quantization are commonly used, under the assumption that the original model need
not be restored (i.e., it is destructive).


The primary goal for this work is to explore compression strategies for
large vector embedding models used in NLP and related use cases, such that one can uncompress and recover the original
model with minimum loss of accuracy.
The \emph{database embedding (db2Vec)}, a
vector embedding technique designed to develop semantic models from multi-modal relational database tables
(\citet{bordawekar:corr-abs-1603-07185, Bordawekar:deem17}), forms
the impetus behind this exploration.  Although related, db2Vec differs from its NLP counterparts such as
Word2Vec~(\citet{mikolov:corr-abs-1301-3781}) and GloVe~(\cite{pennington:glove14}), in that its source
data follows the relational data model~(\citet{DBLP:journals/sigmod/Date82})
(the source data is not a natural language document but a relational database table). Considering the relational database tables
can be very large (e.g., billions of rows in a table) with a large number of unique tokens, a much larger vocabulary is used
than a traditional natural language document. As a result, trained
db2Vec models have high demand on memory capacity and bandwidth (i.e., both memory and AI accelerator I/O bandwidth).

A trained vector
embedding model is a snapshot of it's weight matrices and consists of weight values represented
in IEEE 32-bit single-precision floating point (FP32) format. Therefore, we
focus on compression approaches that compress FP32 to low-precision floating point formats.
In our approach, active rows of the compressed db2Vec model are uncompressed on-demand during inference computations and the original
FP32 values are restored with smaller loss of accuracy than other low-precision methods, as we quantify in later sections.

\begin{figure*}
  \centering
  \includegraphics[width=0.9\textwidth, trim=15mm 145mm 0mm 35mm, clip]{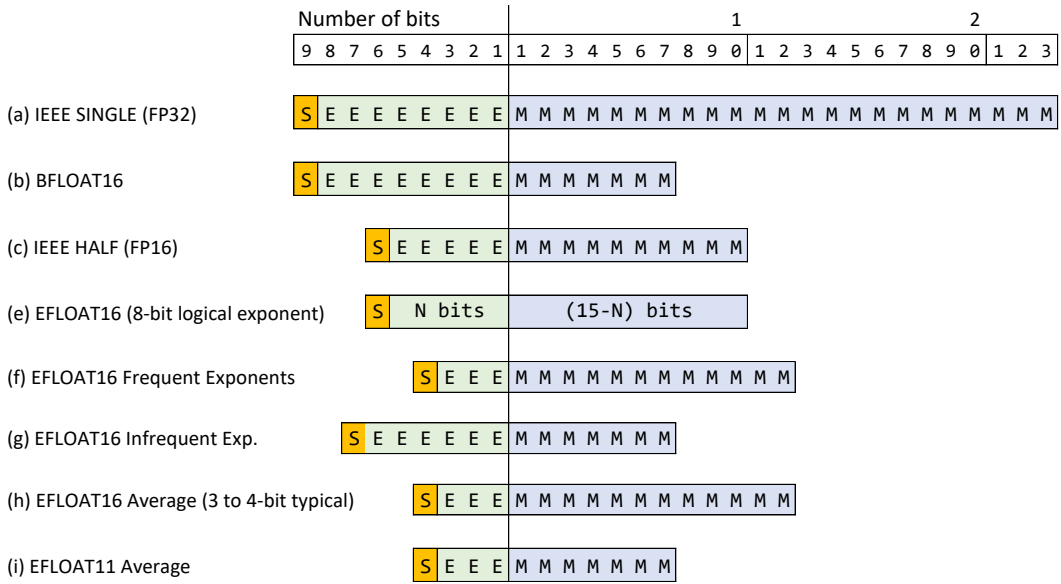}
  \caption{Floating point formats are compared.  EFloat has a fixed total width, but the boundary between the exponent and the significand is variable (e).
    The exponent is entropy coded, providing an average of 4.3 extra bits of precision to the significand (e.g., (h)),
    while keeping the logical exponent range at 8 bits, same as that of FP32.
    EFloat has greater precision and range than the existing FP formats having the same bit budget.
    }
  \label{fig:efloatformat}
\end{figure*}

A floating-point(FP) number is of
the form~(\citet{goldberg:fp}):
\[ -1^{signbit} \times 2^{exponent - bias} \times significand\]
The exponent largely determines the range of minimum and maximum values
representable by the format. The number of significand bits determine the precision
(a constant bias is typically present to represent exponents as all positive
integers which simplifies FP magnitude comparisons.) The two most widely used
low-precision 16-bit FP formats, BFloat16
(\textbf{BF16}) and IEEE 754-2019 Half-precision (\textbf{FP16}),
make a tradeoff between the number of exponent and significand
bits (Figures~\ref{fig:efloatformat}(a,b,c)). BF16 with an 8-bit exponent
and 7-bit significand has a wide exponent range but low precision as
compared to FP32 and FP16~(\citet{wang:bfloat16}).
In contrast, FP16 with a 5-bit
exponent and 10-bit significand has a greater precision but in a much narrower
exponent range than BF16 due to smaller number of exponent bits~(\citet{ieee754}). One could convert a FP32
model to either of these two 16-bit representations, but with a big loss of either accuracy or range,
and the data may not be recovered when converted back to FP32.


Therefore, we have developed a \emph{flexible} and \emph{portable} low-precision compressed FP format, \emph{EFloat
(EFn)}, that can uncompress and recover the original data with minimum loss for runtime inference computations. Figure~\ref{fig:efloatformat}(e) presents the new low-precision FP format, \emph{EFloat
(EFn)}, with a fixed total bit budget of $n$ bits, e.g., $n=16$.
EF16 uses an entropy-coded variable-width $N$ bit exponent and a variable-width $15-N$ bit significand (mantissa) with
a total number of 16 bits, including the sign bit. The EF width is adjustable, e.g., EF12, permitting a tradeoff between
compression ratio and accuracy, as we demonstrate in later sections. We have used the EF format to compress db2Vec models such that relevant rows of the
compressed db2Vec model can then be uncompressed on-demand during inference computations into any target
representation (e.g., FP32 or a 16-bit format) with smaller loss of accuracy.

Our design is motivated by a key
pattern that we observed across a wide range of vector embedding
models: the trained models contain only few of the
$2^{8}=256$ unique exponents available in FP32 and with a bell-shaped histogram distribution
caused by a certain class of non-linear activation functions used in training.
The EFloat design exploits this behavior and assigns
the least number of bits to most common exponent values,
however preserving the original exponent range of the original FP32 format.

Our work makes the following contributions:
\begin{itemize}

\item EFloat provides \textbf{flexible} \emph{variable-length} reduced-bit representation of \emph{any}
  floating point format (e.g., FP32, FP16) by using \textbf{fewer}
  exponent bits to map the \textbf{same} exponent range as the
  original value.

\item For a given bit budget (e.g., 16), EFloat provides more accurate
  representation of the FP32 values than BF16 and FP16 by using fewer exponent bits to capture
  the same range as before, and then using the remaining bits to increase
  significand precision. Thus, EFloat provide \textbf{portable}
  low-precision representation of FP values that can be converted to
  other representations with minimum loss (e.g., 16-bit EFloat
  (\textbf{EF16}) to FP32, BF16, or FP16).

\item We employ the length-limited variant of the Huffman encoding approach
  to map the exponent values to bit codes that satisfy the maximum bit length constraint.

\item Although designed specifically for improving both memory
  and bandwidth via \textbf{compression}, the EFloat format can also support
  native reduced-bit \textbf{computations} over \emph{pre-trained} vector
  embedding models. EFn support in hardware, if required, is basically via 256-entry table lookup converting
  from EFn to conventional FP and vice versa.

\item For a given dataset, multiple floating point to EFloat conversion tables are possible. Tables
  may be optimized for maximum significand width (highest exponent compression) at the expense of
  worse significand precision for few floats with infrequent exponents (i.e., using less significand bits for outliers) and vice versa.

\item Since vector embedding models are used in a wide array of NLP
  architectures including transformers, in addition to db2Vec, the EFloat format can be used for
  a much wider (and more space consuming) class of NLP models.

\item A hardware implementation of EFloat supporting both inferencing and training is sketched.
  EFloat to/from FP conversion tables may be used in the memory and I/O interfaces of AI
  accelerators to save memory capacity and I/O banwidth resulting in higher AI accelerator performance.
\end{itemize}


In Section~\ref{sec:nlpmodels}, we first present the analysis of various vector embedding models.
The EFloat format is presented in Section~\ref{sec:efloatformat}.
Section~\ref{sec:impl} describes key steps in conversion between EFloat and other floating point formats.
Section~\ref{sec:erroranalysis} presents an error analysis of various EFloat widths (EFn) against BF16 and
FP16. In Section~\ref{sec:related}, a review of related work on model compression and floating-point formats
for deep learning is presented. Finally, Section~\ref{sec:concl}
presents conclusions and outlines future directions.

%% file: design.tex
\section{Analyzing vector embedding models}
\label{sec:nlpmodels}

\begin{figure*}
  \centering
  \includegraphics[width=0.8\textwidth, trim=19mm 86mm 19mm 85mm, clip]{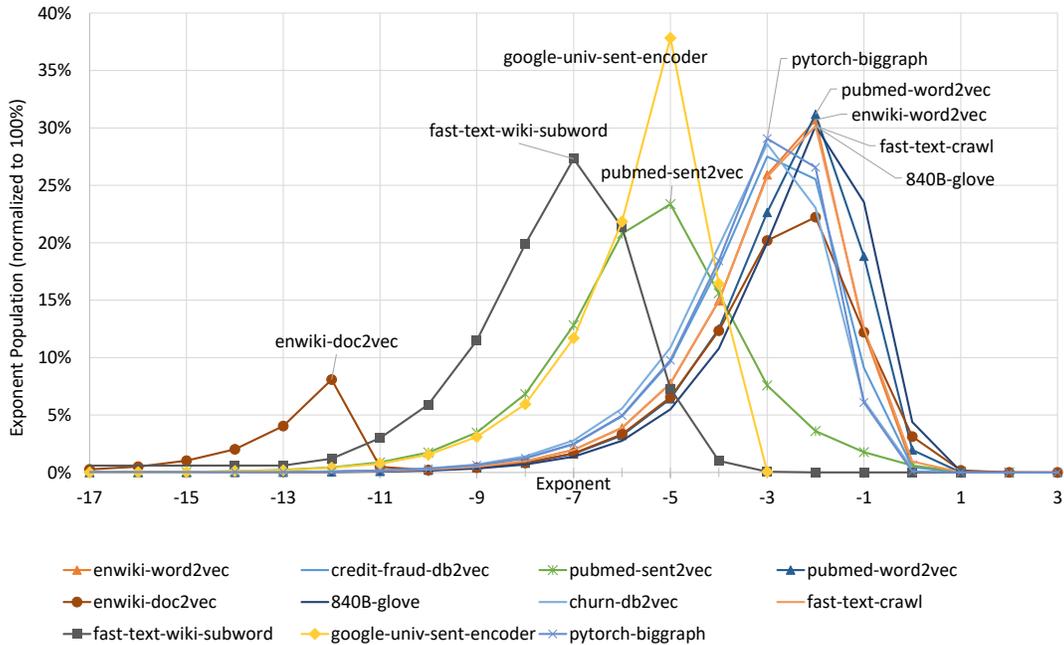}
  \caption{Histogram of the exponent fields of 32-bit floating-point (FP32) values found in vector-embedding and related NLP models.
    Only the db2Vec, word2vec, doc2vec, and sentence-encoder models were generated. Others were downloaded as publically available pretrained models.}
  \label{fig:histogram}
\end{figure*}

Vector embedding models are extensively used in natural language
processing (NLP) to capture and exploit semantic relationships of word entities
(e.g., words, sentences, phrases, paragraphs, or documents). A trained vector
 embedding model consists of a set of vectors, each vector encoding a \emph{distributed}
representation of inferred semantics of a word entity,
i.e., a single vector captures different attributes of the inferred
semantics~(\cite{hinton-distr}), created in part by contributions by other word
entities. Every vector embedding model implements some variant of the
\emph{log-bilinear} language (LBL) model that predicts the probability
of the next word $w_i$ given the previous words
(\emph{context})~(\cite{hinton-slides, almeida2019word, bender-koller-2020-climbing}).  The
LBL model first predicts a real-valued vector representation of a word
by \emph{linearly} combining the real-valued vector representations of
its context words. Then, the distributed representation of the word is
computed based on the similarity between the predicted representation
and the representations of all words in the vocabulary. This step is
accomplished using the \emph{normalized exponential} or \emph{Softmax}
function over the associated feature vectors. The output of the
Softmax function is the probability distribution over $V$ different
possible outcomes, where $V$ is the vocabulary size.

\begin{figure}[ht]
  \centering
  \includegraphics[width=0.48\textwidth]{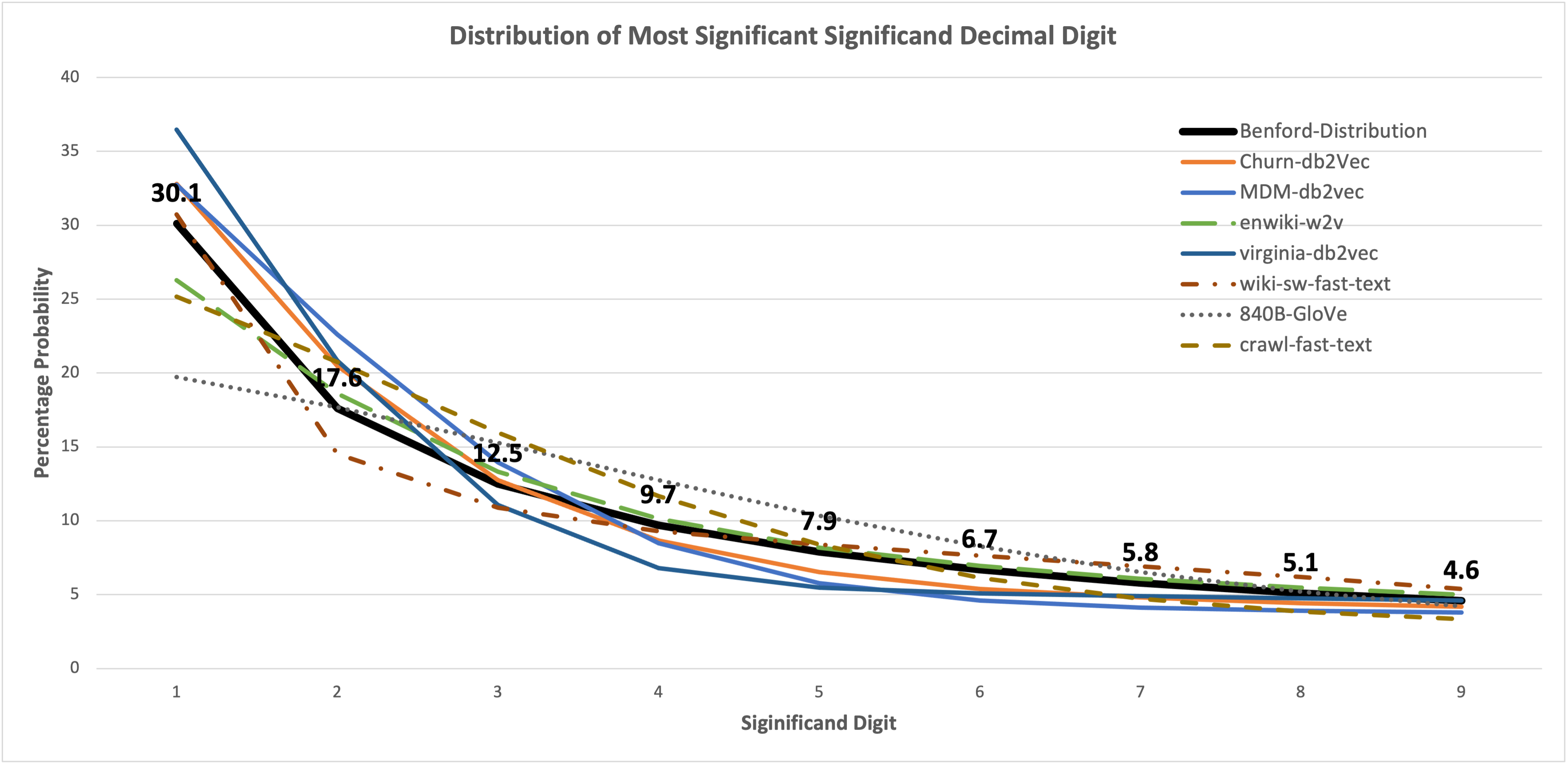}
  \caption{The most significant decimal digit in the significand of FP32 values in various vector embedding models follows the Benford distribution.}
  \label{fig:benford}
\end{figure}

\begin{figure}[ht]
  \centering
  \includegraphics[width=0.48\textwidth]{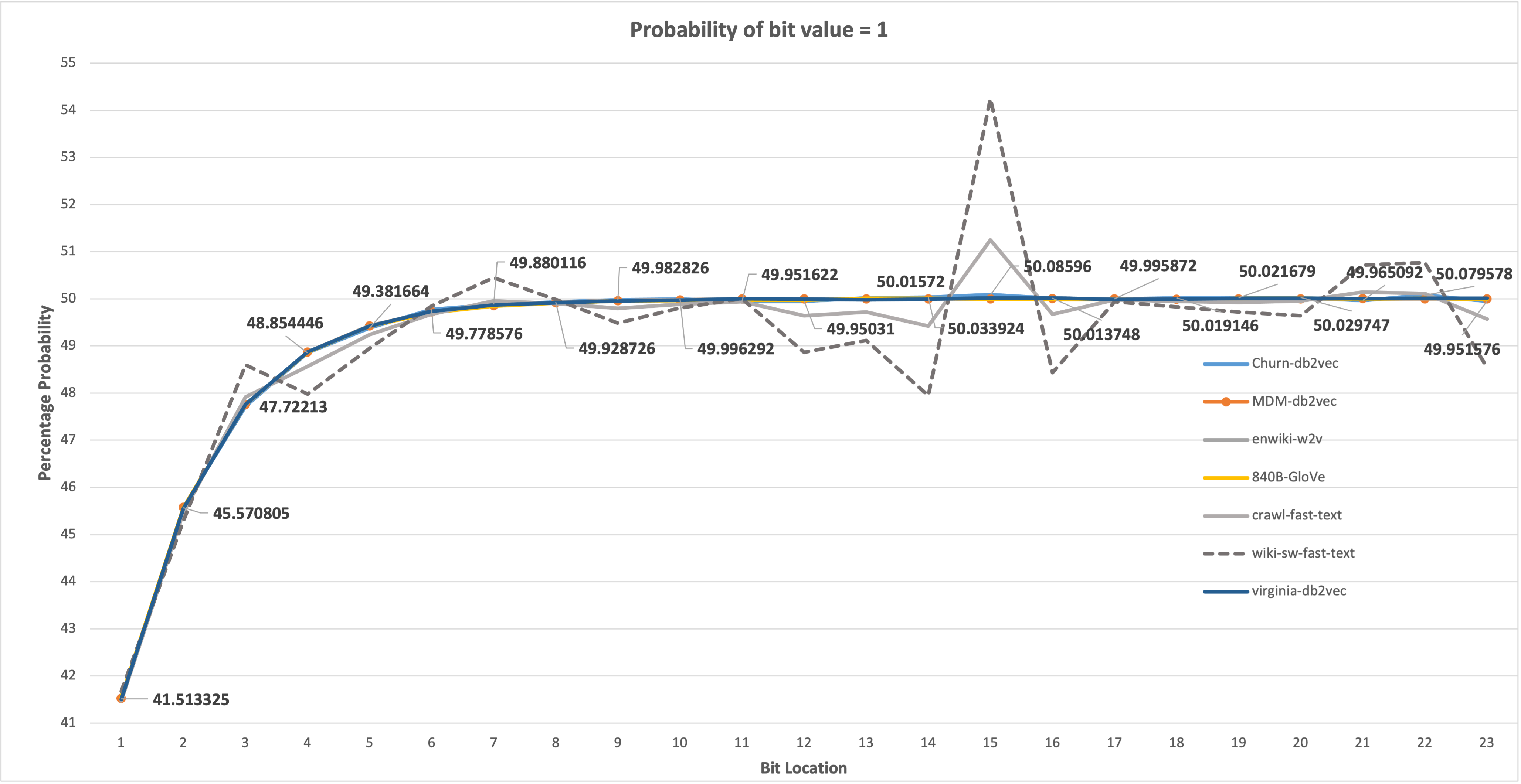}
  \caption{Distribution of a bit value being 1 at different bit locations in the significand of FP32 values in various vector embedding models.}
  \label{fig:bit-distr}
\end{figure}

\textbf{Characterizing exponent behavior:} Figure~\ref{fig:histogram} presents
histograms of exponent values in multiple pre-trained vector embedding
models, where the X-axis represents exponent values (from the 8-bit
exponent portion of a 32-bit IEEE 754 floating point value).
The Y-axis represents normalized number of occurrences of
the exponent values, i.e., a histogram.
For all models, exponents mostly vary between -17 to 3 (i.e., $2^{-17}$ to $2^{3}$).
The most frequent two exponents are -2 and -3 as apparent on Fig.\ref{fig:histogram} except for few models.

 Vector embedding and related NLP models presented in
Fig.~\ref{fig:histogram} include word embedding (word2Vec)~(\citet{mikolov:corr-abs-1301-3781, bio-w2v}),
sentence (sent2Vec) and document embedding (doc2Vec)~(\citet{le:corr14,bio-sent2vec}), GloVe~(\citet{glove-data}), subword embedding (FastText)~(\citet{fast-text,fast-text-data}),
database embedding (db2Vec)~(\citet{bordawekar:corr-abs-1603-07185, Bordawekar:deem17}), graph embedding (PyTorch BigGraph)~(\citet{lerer2019pytorchbiggraph}),
and Google's transformer-based universal sentence encoder~(\citet{cer-etal-2018-universal,tf-encoder}) using the Brown corpus~\cite{brown-corpus}.
All these models implement different
variations of the LBL model. The word2Vec based models, e.g.,
word2Vec, sent2Vec, doc2Vec, db2Vec, and FastText, use a
neural network with different versions of Softmax as the activation
function. GloVe, on the other hand, is a count-based optimization approach that
uses a word co-occurrence matrix and weighted least-square as the
optimization function. The FastText subword model~(\cite{joulin2016bag, fast-text}) assigns a vector
for every character \textit{n}-gram, using an extended skip-gram model~(\cite{mikolov:corr-abs-1301-3781}) and then, words are represented as the sum of these representations. The universal sentence encoder
generates embedding vectors for sentences using a standard Transformer architecture that takes word
embedding vectors as input and uses a Softmax function to compute attention~(\cite{vaswani2017attention}).

Irrespective of the model type, we
observe in Fig.~\ref{fig:histogram} that exponent values cluster in a narrow range of values,
and display a distinct histogram peak. The only exception is
the doc2Vec model that exhibits two peaks as the doc2Vec
first builds fine-grained embeddings for words and then uses them to
build embeddings for coarser-grained entities such as paragraphs via
concatenating and averaging individual word vectors which results in
a smaller second peak. Finally, multiple models of a given type (e.g., churn and credit-fraud db2Vec models), exhibit
the same exponent bahavior.

\textbf{Characterizing significand behavior:} Figures~\ref{fig:benford} and ~\ref{fig:bit-distr} present patterns observed in the significands of FP32 values of
various pre-trained vector embedding models used in the evaluation of the EFloat format. Figure~\ref{fig:benford} plots the frequency
distribution of the most significant significand digit in the decimal representation of the FP32 values. Surprisingly, occurrences of
the digits in these \emph{trained models} follow a logarithmically decreasing distribution consistent with the \emph{Benford's Law}~(\citet{newcomb-law, benford-law}).
Benford's Law predicts the frequency of numbers 1-9 in the leading non-zero digit of most datasets found in nature. According to the law,
the number 1 occurs in the most significant digit 30.1\% of time, the number 2 occurs 17.6\% of time, and so on. The logarithmic reduction
leads to the number 9 being at the most significant location around 4.6\% of time.

Figure~\ref{fig:bit-distr} presents the binary (0/1 bit) view of the significand in few trained models. These models also
approximately follow the Benford's Law for binary numbers. We extended Benford's Law to bits 1 through 23 of the significand.
(Note that the leading non-zero bit in a binary number is always 1 obviously; but that is stored in the hidden leading bit called the \emph{implied-bit}
of normalized FP values, not visible in Figure~\ref{fig:bit-distr}.)
In this figure, we report the agregate distribution of a particular FP32 significand bit location having the value 1,
for the 23 significand bits in selected vector embedding models. As we observe, the distribution exhibits
a knee at location 7; the first 7 bits show an increase in the probability of the bit value being 1 from 41.51\% to 49.88\%;
the remaining bits exhibit probabilities around 50\%. In other words, it is more likely to see the bit value being 0 in the most significant bit
of an FP32 significand in trained vector embedding models.  That the bit values in trained models are skewed towards 0 in the leading bits presents additional
compression and FP rounding-mode opportunities to improve precision, although the focus of this paper is mainly on the FP exponent.

\begin{figure}[ht]
\centering
\includegraphics[width=0.48\textwidth, trim=19mm 97mm 19mm 95mm, clip]{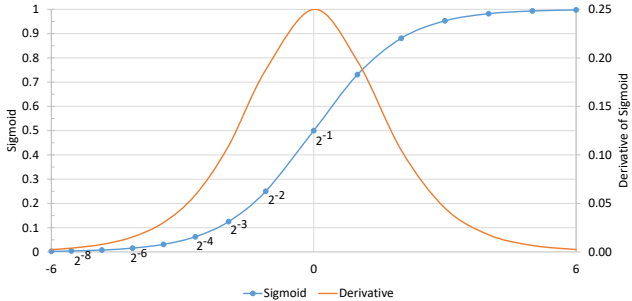}
\vskip -1mm
\caption{The Sigmoid $\sigma(x)$ curve and its gradient. The floating-point (FP32) exponent of few neural weights are overlaid on $\sigma(x)$.}
\label{fig:sigmoid}
\end{figure}

\begin{figure}
\centering
\includegraphics[width=0.48\textwidth, trim=19mm 99mm 19mm 95mm, clip]{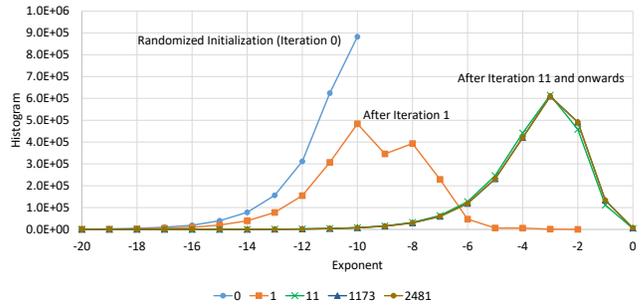}
\vskip 0mm
\caption{Exponent population (FP32) of model weights during
the training of churn-db2vec model, for iterations 0 to through 2481}
\label{fig:churn}
\end{figure}

\textbf{Impact of activation function:} The Softmax family of activations functions used in vector embedding models is responsible for
the clustering behavior of exponents (Figure~\ref{fig:histogram}). To understand the reasons, let
us delve deeper into the training of an embedding model. For illustration purposes, we use database
embedding (db2Vec) of the Telecom Churn data~(\citet{churn-data}) as an example. db2Vec is an adaptation of the word2Vec approach, and has been
designed to build an embedding model from structured database tables that adhere to the relational data model. Like
word2Vec, db2Vec also uses Skipgram with Negative Sampling (SGNS) as the training approach. The SGNS approach uses a binary
classifier based on the logistic (Sigmoid) function instead of using the Softmax-based predictor. The Sigmoid function
$\sigma(\boldsymbol{x})$ is defined as:

\[ \sigma(\boldsymbol{x}) = \frac{1}{1 + e^{-x}} \]

The Sigmoid curve and it's gradient $\sigma'(x) = \sigma(x)(1-\sigma(x))$ is shown in Figure~\ref{fig:sigmoid}.
The gradient of the Sigmoid activation function and the back-propagation process can be specified as:
\[ \text{\it weights}_{\text{iter}+1} = \text{\it weights}_{\text{iter}} + \text{\it learning\_rate} * \sigma'(x) \]

\begin{figure}
  \centering
  \includegraphics[width=0.48\textwidth, trim=19mm 95mm 19mm 94mm, clip]{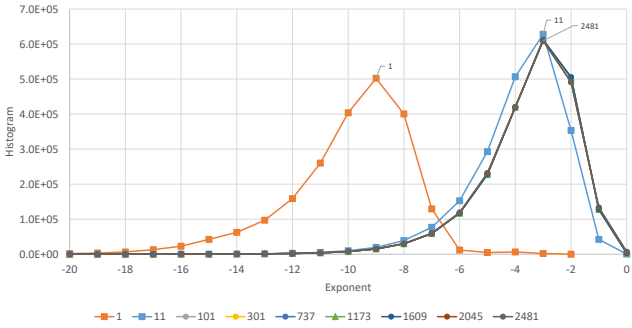}
  \caption{Population of floating-point (FP32) exponents of update values ($learning\_rate*\sigma'(x)$)
during churn-db2vec model training for iterations 1 to through 2481}
  \label{fig:churngrad}
\end{figure}

\begin{figure}
  \includegraphics[width=0.48\textwidth, trim=19mm 95mm 19mm 104mm, clip]{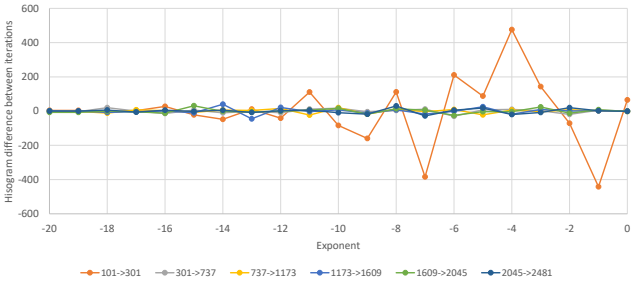}
  \caption{Differences in the floating-point (FP32) exponent populations between certain iterations during training of the churn-db2vec model}
  \label{fig:churniterationsdelta}
\end{figure}

The overall training process involves multiple back-propagation iterations to update model weights using the gradient
of the Sigmoid function. Weights get updated iteratively during the back-propagation process by
the error computed for that iteration. Practically, the error is
computed using the gradient of the activation function.
During model training, we observe that the weights rapidly converge~(Fig.\ref{fig:churn}) to
their final values. Their exponents are substantially clustered at the slope of the Sigmoid curve,
the $2^{-8}$ to $2^{0}$ output range of Sigmoid, as evidenced by Figures~\ref{fig:sigmoid} and~\ref{fig:churn}.
Training eliminates smaller exponents from the model because the activation function
output is practically zero for any input value when weights are small.
Large exponents are non-existent due to normalization of weights. Accordingly, most exponents cluster at the slope of the Sigmoid curve.
Further, we
observe that early in the
training process, the significand precision is not as important as
the magnitude of model weight updates are dominated by exponent updates from one iteration to
the next (Figures~\ref{fig:churngrad} and ~\ref{fig:churniterationsdelta}).  Once an exponent settled to its final value,
the significand precision becomes more important since weight starts converging
to its final value in small increments.

%
%
%

%

This observation is useful for amortizing the overhead of Huffman
code-table generation over many training iterations and inferencing.
One could produce a static code-table covering multiple NLP models (except a few models
such as Wiki-doc2vec) shown in
Figure~\ref{fig:histogram}. Then, one could use the same code-table
any number times because the exponent frequencies hence the
code-tables are practically identical across these models.  A pre-built
code-table may be optimized for the
final iteration of training but the used for all iterations. That the pre-built code table is not optimized
for early training iterations is not important.
Because early in the training, exponents are changing rapidly and
the significand precision is not as important.


%% file: efloat.tex
\section{The EFloat format: EF$n$}
\label{sec:efloatformat}


The key idea behind the EFloat format is the variable-width encoding of exponents using
the well-known Huffman algorithm.
Frequency of unique exponent values in the dataset determine
the coded-exponent widths which
may vary between as small as 1-bit and some software configurable maximum, e.g., 8-bit (Figure~\ref{fig:efloatformat}(e,f,g)).
Thanks to the entropy coding of the Huffman algorithm, frequent exponent values are coded with fewer bits
and infrequent exponents are coded with more bits as observed in Fig.\ref{fig:enwiki1}.
For example, the most frequent exponent -2 is encoded with 2 bits. That shaves 6-bits out of
the 8-bit exponent field used in the conventional FP32 format.




Bits saved from the exponent
become part of the significand, therefore increasing the floating point precision compared to other float formats with the same bit budget.
An $N$-bit coded-exponent in an EF16
float results in a (15-$N$) bit significand as shown in Fig.\ref{fig:efloatformat}(e).
Since EFloats with frequent exponents have wide significands, the entire dataset has a greater precision on average.
EFloats with infrequent exponents have narrow significands. But, their contribution are relatively small in the common
calculations used in model training and inferencing, such as dot-products, vector-sums, and cosine-similarity
(EFloat precision is quantified and compared to prior formats in Section~\ref{sec:erroranalysis}.).

EFloat on average have greater precision and range than any other fixed-field FP
format with the same bit budget.  For example, EF16 with a
3-bit coded-exponent has 12-bits of significand compared to the 7-bit significand
found in a BF16 (Figures~\ref{fig:efloatformat}(h,b)).
However, EFloat exponent's logical width is \emph{always} 8-bit, the same
as for FP32 and BF16, irrespective of EFloat width.
Even for extremely narrow floats such as EF8, the logical exponent width can
be 8-bit since encoding compresses the exponent field.

The EFloat format compresses special
values of IEEE 754, such as signed zeros and infinities
losslessly.  NaN are semantically compressed losslessly: converting a NaN
to and from FP32 to EFn and vice-versa still results in a NaN.
Denormal floats may round to zero since least significant bits of significands are truncated
during encoding.

%% file: impl.tex
\section{EFloat encoding and decoding}
\label{sec:impl}

\subsection{Entropy Compression}

\begin{figure}[ht]
\centering
\includegraphics[width=0.48\textwidth, trim=19mm 107mm 19mm 110mm, clip]{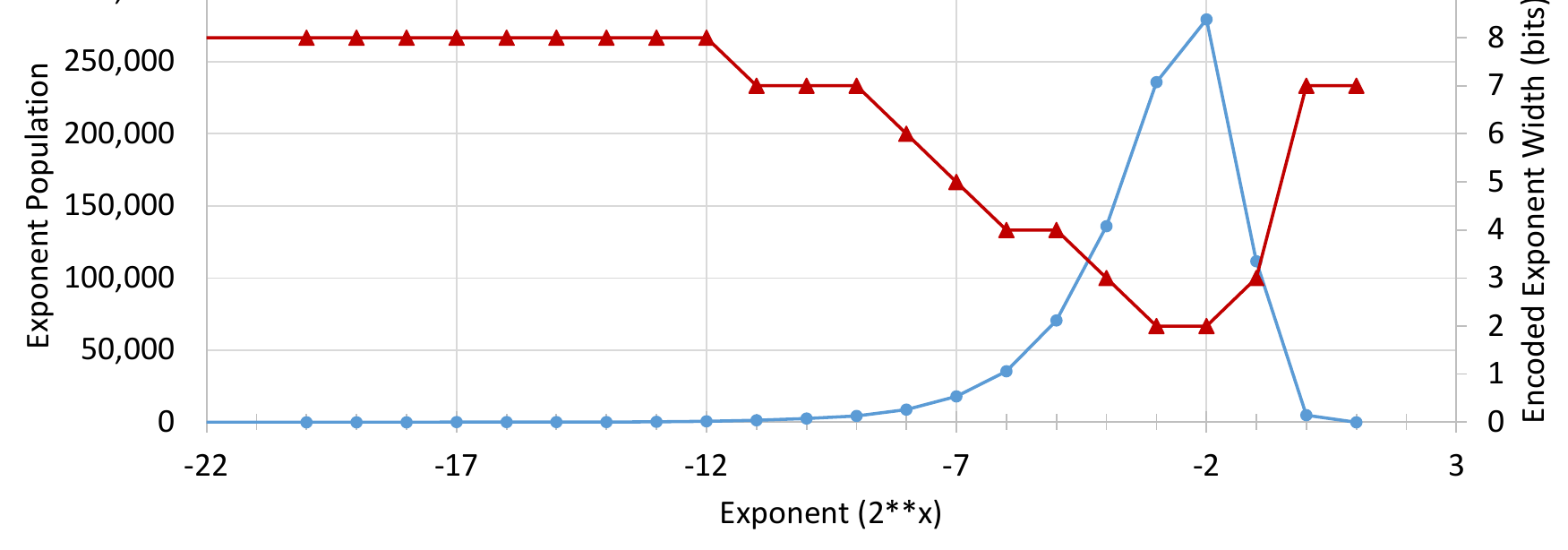}
\caption{EFloat variable exponent widths are a function of the exponent population (enwiki-word2vec)}
\label{fig:enwiki1}
\end{figure}

\textbf{The Huffman algorithm:} is a popular lossless compression
algorithm used in many compression tools and compressed data
formats~(\citet{salomon2004data}). Data symbols are encoded with
variable-length binary codes whose length are determined by the symbol
probabilities in the data stream.  The algorithm builds a binary tree
with each leaf assigned a symbol.  Higher probability symbols are
closer to the tree root than others.  The path from the tree root to
the leaf is the binary coding of the symbol.  To demonstrate with a
trivial example, the letters A, B, C occuring with probabilities of
0.5, 0.25, and 0.25 may be encoded with the bit patterns 0, 10, and
11, respectively.  The algorithm yields 1.5-bit/symbol compression
efficiency, better than 8-bits/symbol using an ASCII representation or
2-bits/symbol using a simplistic mapping of the 3 letters to 2-bit
integers. Huffman coding is optimal when symbol probabilities are
negative powers of 2. However, it is an effective compression method even for non powers of 2
distributions. Fig.\ref{fig:enwiki1} shows the Huffman coded exponent
widths as a function of exponent frequencies of a word2vec trained model.

Huffman codes have the {\em prefix} property that states that no
code is a prefix of a longer code (due to the binary tree construction.) As a
result, the Huffman code not only encodes the original symbol but the
code-length as well.  Thanks to the prefix property,
the movable boundary between the exponent and the significand is easily
identified while decoding EFloats(~Fig.\ref{fig:efloatformat}(e)).


\textbf{Length-limited Huffman Encoding:} The Huffman
algorithm builds probabilities based on frequency histogram of exponents and
outputs a code-table mapping 8-bit exponents to variable-width
coded-exponents.

If the dataset contains
$N$ unique exponents, for some distributions the algorithm may
produce codes with $N-1$-bit widths in the worst case exceeding EFloat's entire
width.  For example, a word2vec dataset in
Fig.\ref{fig:enwiki1} contains 23 unique exponent values.  With a
worst case distribution, some coded exponent widths
may not fit in an EF16 number (e.g., a 22 bit exponent). Therefore, we use the {\em Length-Limiting} variant
of the Huffman algorithm to set a maximum coded-exponent width~(\citet{abali2020data})\footnote{Code
for the length-limited Huffman Encoding is available at https://github.com/libnxz/power-gzip/blob/master/lib/nx\_dhtgen.c}.
In Fig.\ref{fig:enwiki1}, the maximum
code width is set to 8-bits resulting in the infrequent exponents coded with that maximum.
Note that length-limited Huffman coding is a well know compression technique and has been in use in many popular compression tools.
Our contribution here is application of it in a floating-point number format.


\begin{figure}
\centering
\includegraphics[width=0.48\textwidth, trim=19mm 107mm 19mm 110mm, clip]{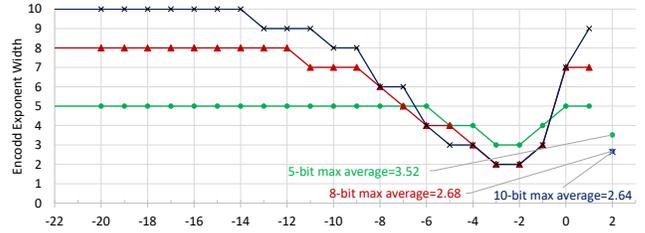}
\vspace{0mm}
\caption{EFloat uses Length-Limited Huffman algorithm to set the maximum width of coded-exponents to 5, 8, or 10-bits}
\label{fig:lengthlimited}
\end{figure}



This software-defined maximum exponent width presents an opportunity to tune the EFloat
precision to the particular NLP application requirements.
Figure~\ref{fig:lengthlimited} shows the inverse relationship between the maximum
code-width and the minimum code-width. As the limit is increased from 5 to 8 and 10-bits
we observe that the least-frequent exponents are coded with the maximum-width
codes as expected, therefore their respective significands lose precision.
At the same time, with an increased limit the most-frequent exponents are now coded with fewer bits.
This reduces the average coded-exponent width as Figure~\ref{fig:lengthlimited} shows. Therefore,
on average the floating point values gain precision while few outliers lose precision.  Therefore,
EFloat not only compresses the regular floats but for a given EF$n$
budget of $n$-bits the software application can optimize the end-to-end precision and compression ratio by
adjusting both the floating point bit-budget $n$ and the maximum coded-exponent width.

\subsection{EFloat Encoding and Decoding}

\textbf{EFloat Encoding and the Code-Table:} During the conversion
from FP32 to an EFn (e.g., EF16), exponents in the original
dataset are histogrammed first, e.g., Fig.\ref{fig:histogram}.
The histogram representing probabilities of the exponents is an
input to Length-Limiting Huffman algorithm that produces a code table
for translating the FP32 exponents
to EFn coded-exponents.
The output is a 256-entry ($2^8$) code-table
indexed by the original 8-bit exponent.
Each table entry contains a pair, the variable-width coded-exponent and its bit-width.
Note that the code-table is quite small, tens of bytes in practice, since few unique exponents
are present in most NLP datasets as Fig.\ref{fig:histogram} shows.

When the sign bits have a skewed
distribution, e.g., if they are substantially positive, then the
sign bit and the 8-bit exponent may be treated as a single 9-bit
integer when histogramming.  Thanks to the entropy coding, a skewed sign bit distribution may
save up to one additional bit in the exponent, becoming available for further increasing precision of the significand.

Using the code-table, the entire dataset is converted from FP32 to the chosen EFn width (e.g., EF16)
replacing original exponents with coded-exponents.  Least
significant bits of the FP32 significand are truncated to match the EFn
width.  For example, in Fig.\ref{fig:enwiki1}, the algorithm
encodes the most frequent exponent with 2-bits. Accounting for the
sign bit, this yields a 13-bit EF16 significand by
truncating the bottom 10-bit of the 23-bit
significand of FP32. We use the {\em round-to-nearest} method
to provide on average 0.5 bits of additional precision during this truncation step
(many different rounding modes may be used, although it's not a subject of this paper.)

For large datasets, a statistically representative subset may also be
used to reduce histogram collection time. Furthermore, when the histogram is known
in advance, a pre-built code-table may be used, skipping the histogram collection and Huffman algorithm steps.
During training exponents rapidly converge to their final values as observed
in Fig.\ref{fig:churn}.  Past iteration 10, the exponent distribution is practically
identical for all iterations 11 through 2481, which suggests that a single pre-built
code-table optimized for final iterations may serve for all iterations
start to finish. The same pre-built table, although suboptimal for early iterations, may be used
because significand precision is not as important at that point in time; model weight updates are dominated by
exponent updates.  Once exponents settled to their
final values the significand precision becomes important since model
weights updates progressively get smaller.

\textbf{EFloat Decoding:} For EFloat to FP32 conversion we use a inverse mapping of the code-table
described earlier.
Note that Huffman codes are {\em prefix} codes which encode both the
original value and the code-width.  Therefore, the movable boundary
between the exponent and the significand is not ambiguous; a
``boundary marker'' is not necessary.
A decoder-table indexed by the coded-exponent
may be used to decode the original exponent value and the significand's
leading bit position in constant time. The decoder-table has as many entries as $2^{max\_code\_width}$
e.g., $2^{8}$.  Each table entry contains the original
exponent and bit-width of the coded-exponent.  To index the decoder-table with variable-width codes
many entries are filled with duplicates. For example,
a 2-bit coded-exponent $00$ is duplicated 64 times in the table
at locations ${\bf 00}000000$ through ${\bf 00}111111$ with each
location containing the original exponent and code-width$=2$.
Duplicating entries is equivalent
to having {\em logical don't care} bits in the index which is a common technique used in hardware lookup tables.

The second element of each table entry contains the EFloat significand width.
Since the significand was truncated earlier during the FP32
to EFloat conversion, the missing least significant bits must be
padded with zeros to match the original FP32 width.

\subsection{Implementation Alternatives} Currently, both EFloat encoding
and decoding are completely implemented in software that allows us to
compress an FP32 value to an EFn format of variable size $n$, and conversely, given a value
in the EFn format, generate its FP32, BF16, or FP16
representations.

As hardware arithmetic functional units natively
supporting EFloat operations currently do not exist, it is necessary
to recover data from the compressed EFloat representation into the
target floating point representation used in the computation.
EFloat encoding is not time critical as it's a trivial fraction of training time.
However,
hardware support for EFloat decoding may be necessary so that the numerical values in
the correct format are fed to the functional units with minimum
delay. The EFloat decoding hardware may be implemented with a Static Random Access Memory (SRAM) based lookup table.
With a maximum code width of $K$ bits, a $2^K$ entry SRAM based table
may be used. $K=8$ is desirable to cover the worst case condition of
all 256 exponents utilized by an FP32 dataset.

If the compute processor has many input ports, for example a systolic
array such as the Google TPU~\cite{jouppi2018motivation}, or a wide
SIMD architecture~\cite{henry2020high,hennessy2017computer}, many
decoder tables will be necessary for
parallel access.  To save area, the maximum code width may be set to
$K < 8$ at the expense of giving up some compression quality.  An
alternative approach for saving area might be using a two level table
which will substantially reduce SRAM capacity requirements.
For example, the Zlib
software~\cite{zlib:library} uses a maximum code width of 15-bits.  But
for encoding the Zlib symbols, a 2-level table requires only 852
table entries compared to $2^{15}=32768$ entries required in a 1-level table.


%% file: eval.tex
\section{Evaluating the EFloat representation}
\label{sec:erroranalysis}

In this section, we evaluate the efficacy of the EFloat format using two sets
of experiments. The first set measures the loss of precision in representing FP32 data in BF16, FP16,
and EFloat formats with bit budgets from 16 down to 8 bits. The second set of experiments
compares the quality of ranked results for \emph{similarity} and \emph{dissimilarity} queries using the \emph{Normalized Discounted
Cumulative Gain (NDCG)} score for BF16, FP16, and various EFloat
formulations.  Table~\ref{tab:datasets} presents the
list of the pre-trained vector embedding models used in these experiments, along with their
characteristics: mode types, number of tokens in the model, the vector
dimension and the model size (vector values are stored using FP32).

\begin{table}[htbp]
\caption{Vector embedding datasets used for evaluation}
\centering
\begin{tabular}{llrcr}
\toprule
 Name    &  Type     & \#Tokens & Dimension & Size \\
\cmidrule(lr){1-5}
churn    & db2vec    & 7104     & 300       & 20 MB  \\
crawl    & fast-text & 1999995  & 300       & 4.3 GB \\
enwiki   & word2vec  & 5427849  & 200       & 9.6 GB \\
MDM      & db2vec    & 5050264  & 300       & 14 GB  \\
840B     & GloVe     & 2196017  & 300       & 5.3 GB \\
wiki-sw  & fast-text & 999994   & 300       & 2.2 GB \\
virginia & db2Vec    & 80772    & 300       & 222 MB \\
\bottomrule
\end{tabular}
\label{tab:datasets}
\end{table}

Table~\ref{tab:exp-char} presents characteristics for the models used
in these experiments: code table sizes, number of unique exponents,
range of exponent bits generated by the Huffman algorithm, the
average count of exponent bits, and minimum and maximum average count of significand bits.
For EF16, the average significand length is 4.3 bits higher than
BF16 (with a 7-bit significand) and 1.2 bits higher than FP16 (with a 10-bit
significand). EF8 is another demonstration of the EFloat benefit:
the full 8-bit exponent range, a sign bit, and a 3.4 bit avg. width significand (a total of 12.4 bits) fit in a budget of 8 bits.
As the Table~\ref{tab:exp-char} illustrates the
code tables used for encoding and decoding these models are very small
in size (average size is 86 bytes and the maximum is 116 bytes).

\begin{table*}[htbp]
\caption{EFloat characteristics from EF16 to EF8 for different datasets}
\centering
\begin{tabular}{lccccccc}
\toprule
\multirow{2}{*}{Model} & Code Table Size & Unique & \multicolumn{3}{c} {EFn exponent bits} & \multicolumn{2}{c} {EFn significand bits (Avg.)} \\
    &  Bytes & exponents              &     Min      & Max        & Avg.      &  Max (EF16) & Min (EF8) \\
\cmidrule(lr){4-6} \cmidrule(lr){7-8}
churn & 80 & 23 & 3  & 5 & 3.6 & 11.4 & 3.4 \\
crawl & 71 & 30 & 3 & 5 & 3.4 & 11.6 & 3.6 \\
enwiki & 92 & 27 & 4 & 5 & 4.2 & 10.8 & 4.8 \\
MDM  & 83 & 24 & 3 & 6 & 3.6 & 11.4 & 3.4 \\
840B & 116 & 35 & 3 & 6 & 3.5 & 11.5 & 3.5 \\
wiki-sw & 77 & 22 & 3 & 5 & 3.6 & 10.5 & 3.4 \\
virginia & 83 & 24 & 3 & 5 & 3.7 & 11.3 & 3.4\\
\bottomrule
\end{tabular}
\label{tab:exp-char}
\end{table*}

\subsection{Evaluation of Numerical Precision}

The first set of experiments compares the loss of precision due to the least significant significand
bits being truncated during conversion from FP32 to various lower-precision formats. Given a low-precision format
(e.g., EF16 or BF16), the values are converted back to FP32, and the arithmetic difference, $f^o - f^c$,
of the original FP32 value, $f^o$, and the regenerated FP32 value, $f^c$, is computed. This difference represents the precision
loss due to conversion. Root Mean Square Error (RMSE) metric is then used to summarize the loss of precision across a
dataset of $N$ floats as: \[ RMSE = \sqrt{\frac{1}{N} \sum_{k}^{N}  (f_{k}^o - f_{k}^c)^2} \]
We then compare the errors of BF16/FP16 and EFn by dividing $RMSE_{BF16/FP16}$ by
$RMSE_{EFn}$ in Table~\ref{tab:errors}. Ratios greater than $1.0$
indicate that the EFloat error is less than BF16 or FP16 errors. For EF16, across all models, we observe an average RMSE error ratio
of 24.1 for BF16, and 1.5 for FP16.  Note that for these experimental results, the datasets
were encoded with a minimum of 3-bit and a maximum of 6-bit coded-exponents
resulting in an average width in the range of 3.4
to 4.2-bits (Table~\ref{tab:exp-char}). Accordingly, for EF16, the \emph{minimum} significand width
is 10-bit which is 3-bit wider than BF16, and of the same length as FP16. Therefore,
EF16 has significantly higher precision against BF16 than FP16.  Also, Table~\ref{tab:errors} shows that
EF12 has the same to slightly better RMSE than BF16 since the RMSE ratios are in the 1.0 to 2.2 range.  Thus,
EF12 uses 25\% less bandwidth and memory capacity than BF16 for
similar floating-point precision.

\begin{table*}

\caption{BFloat16 (BF16), IEEE Half (FP16), and EF16--8 precision
  comparisons using FP32 RMSE ratio. Higher is better.}
\begin{adjustbox}{width=2.1\columnwidth,center}
\centering
\begin{tabular}{lcccccccccccccccccc}
\toprule
& \multicolumn{2}{c}{EF16} & \multicolumn{2}{c}{EF15} & \multicolumn{2}{c}{EF14} & \multicolumn{2}{c}{EF13} & \multicolumn{2}{c}{EF12} & \multicolumn{2}{c}{EF11} & \multicolumn{2}{c}{EF10} & \multicolumn{2}{c}{EF9} & \multicolumn{2}{c}{EF8} \\
& BF16 & FP16 & BF16 & FP16 & BF16 & FP16 & BF16 & FP16 & BF16 & FP16
       & BF16 & FP16 & BF16 & FP16 & BF16 & FP16 & BF16 & FP16\\
\cmidrule(lr){2-3} \cmidrule(lr){4-5} \cmidrule(lr){6-7}
  \cmidrule(lr){8-9} \cmidrule(lr){10-11} \cmidrule(lr){12-13}
  \cmidrule(lr){14-15} \cmidrule(lr){16-17} \cmidrule(lr){18-19}
churn & 22.5 & 1.4 & 11.3 & 0.7 & 5.6 & 0.4 & 2.8 & 0.2 & 1.4 & 0.009 & 0.7 & 0.04 & 0.3 & 0.02 & 0.2 & 0.01 & 0.08 & 0.005 \\
crawl & 34.6 & 2.2 & 17.3 & 1.1 & 8.6 & 0.5 & 4.3 & 0.3 & 2.2 & 0.1 & 1.1 & 0.07 & 0.5 & 0.02 & 0.3 & 0.02 & 0.1 & 0.008 \\
enwiki      & 16.9 & 1.0 & 8.5 & 0.5 & 4.2 & 0.3 & 2.1 & 0.1 & 1.0 & 0.07 & 0.5 & 0.03 & 0.3 & 0.02 & 0.1 & 0.008 & 0.06 & 0.004 \\
MDM      & 27.9 & 1.8 & 13.9 & 0.9 & 6.9 & 0.4 & 3.5 & 0.2 & 1.8 & 0.1 & 0.9 & 0.05 & 0.4 & 0.03 & 0.3 & 0.01 & 0.1 & 0.007 \\
840B           & 25.0 & 1.6 & 12.5 & 0.8 & 6.3 & 0.4 & 3.1 & 0.2 & 1.6 & 0.09 & 0.8 & 0.05 & 0.4 & 0.02 & 0.2 & 0.01 & 0.09 & 0.006 \\
wiki  & 22.0 & 1.4 & 11.02 & 0.7 & 5.5 & 0.3 & 2.7 & 0.2 & 1.2 & 0.08 & 0.6 & 0.04 & 0.3 & 0.02 & 0.2 & 0.01 & 0.08 & 0.005 \\
virginia & 19.6 & 1.2 & 7.8 & 0.6 & 4.9 & 0.3 & 2.4 & 0.2 & 1.2 & 0.08 & 0.6 & 0.04 & 0.3 & 0.02 & 0.2 & 0.009 & 0.07 & 0.004 \\
\bottomrule
\end{tabular}
\end{adjustbox}
\label{tab:errors}
\end{table*}

Table~\ref{tab:errors2} presents the relative RMS error ratios of
BF16, FP16, and EF16 to evaluate the impact of two
different rounding strategies: deterministic (\emph{DETR}) and
stochastic (\emph{STOC}) roundings. In the \textbf{DETR} approch, if the leading bit of
the truncated part is 1 the significand is incremented by 1 (using
integer arithmetic) provided it doesn't overflow in to the exponent
field. If the leading bit is 0, the significand is not
incremented. In \textbf{STOC}, we adapt the stochastic rounding
approach~\cite{HOHFELD1992291, gupta:stoc, zhang2021fast, mikaitis2020stochastic}  as follows: given $n$ bits
to be chopped from the significand, we increment the significand by 1
with probability of $value(n)/(2^{n+1})$ and do not increment with
probability 1-$value(n)/(2^{n+1})$. As Table~\ref{tab:errors2} shows,
for all datasets, the we observed higher RMS error ratios (i.e.,
better accuracy) when deterministic rounding was used. Therefore, we
use deterministic rounding as the default rounding approach.

\begin{table*}
\caption{Impact of deterministic(DETR) and stochastic(STOC) rounding
  on BF16, FP16, and EF16, using the RMSE-with-FP32 ratio. Higher is better.}
\begin{adjustbox}{width=2.1\columnwidth,center}
\centering
\begin{tabular}{lcccccccccccccccc}
\toprule
\multirow{2}{*}{Base} & \multicolumn{2}{c}{churn} & \multicolumn{2}{c}{crawl} & \multicolumn{2}{c}{enwiki} & \multicolumn{2}{c}{MDM}  & \multicolumn{2}{c}{840B} & \multicolumn{2}{c}{wiki-sw} & \multicolumn{2}{c}{virginia}  \\
    & DETR & STOC   & DETR & STOC      & DETR & STOC      & DETR & STOC   & DETR & STOC   & DETR & STOC      & DETR & STOC   \\
\cmidrule(lr){2-3} \cmidrule(lr){4-5} \cmidrule(lr){6-7}
  \cmidrule(lr){8-9} \cmidrule(lr){10-11} \cmidrule(lr){12-13} \cmidrule(lr){14-15}
BF16 & 22.5 & 11.9 & 34.6 & 18.3 & 16.9 & 8.8 & 27.9 & 14.6 & 25.0 & 13.0 & 22.0 & 11.5 & 19.6 & 10.2 \\
FP16 & 1.41 & 0.7 & 2.2 & 1.1 & 1.0 & 0.6 & 1.8 & 0.9  &  1.6 & 0.8 &  1.4  &   0.7 & 1.2 & 0.6 \\
\bottomrule
\end{tabular}
\end{adjustbox}
\label{tab:errors2}
\end{table*}

\subsection{Evaluation of Result Quality}

Note that the RMSE method amplifies larger errors
due to the squaring of differences. EFloat coded floating point values with
short significands (i.e., those with infrequent exponents) are
disproportionately represented in the RMSE summation. However, the true measure of error for
vector embedding models will be the evaluation of ranked results for similarity queries for different
floating point formats. Unlike the binning in traditional classification inference tasks,
ranked results from similarity queries are far more sensitive to numerical precision. We use
the {\em Normalized Discounted Cumulative Gain (NDCG)} metric~(\citet{cg, ndcg}), to evaluate the quality of
ranked results for different floating point formats. NDCG is widely used in information retrieval and web search to evaluate the relevance of retrieved documents.
NDCG is a normalization of the Discounted Cumulative Gain (DCG)
measure. DCG is calculated as a weighted sum of the degree of
relevancy of the ranked items, where the weight is a decreasing
function of the position of an item. NDCG is computed by normalizing
DCG by IDCG, which
is the DCG measure for a perceived ideal ranking result. Thus, the
NDCG measure always lies within [0.0,1.0]. The NDCG metric provides us a common
evaluation format across multiple vector embedding models, irrespective of their target use cases.


For a given vector embedding model, we choose $q=20$ randomly selected distinct query points. For each query point, we compute similar and dissimilar
points by computing cosine similarities over the corresponding vectors. For similarity queries, the result contains a list of points sorted
in decreasing order of their similarity scores (most similar pair of items will have score closer to 1.0), and for dissimilarity queries,
the result list is sorted in increasing order of their similarity scores (most dissimilar pair of items will have score closer to -1.0). For
each query point, we run similarity and dissimilarity queries for different floating point formats,
and use the top $k=10$ results for each test to compute the NDCG
score, (\textbf{NDCG@10}). In our evaluation, we use the ranked
results for FP32 as the baseline for calculating the IDCG. For each
model, we report the average NDCG@10 score computed over 20 query
points using BF16, FP16, and various EFn from EF16 to EF8.

\begin{figure*}[htbp]
  \centering
  \includegraphics[width=0.9\textwidth]{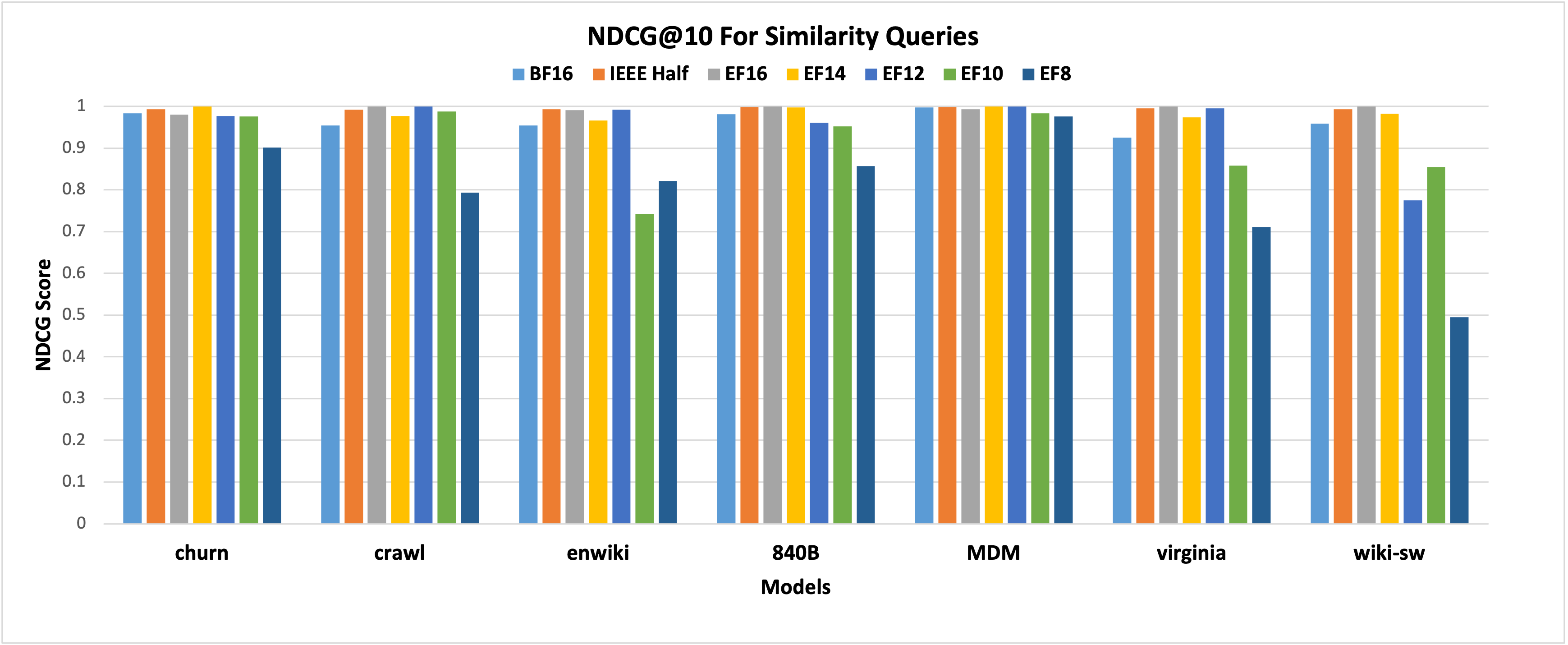}
  \caption{Evaluation of similarity query accuracy using NDCG score across different floating point formats. Higher score (closer to 1.0) is better.}
  \label{fig:sim}
\end{figure*}

\begin{figure*}[htbp]
  \centering
  \includegraphics[width=0.9\textwidth]{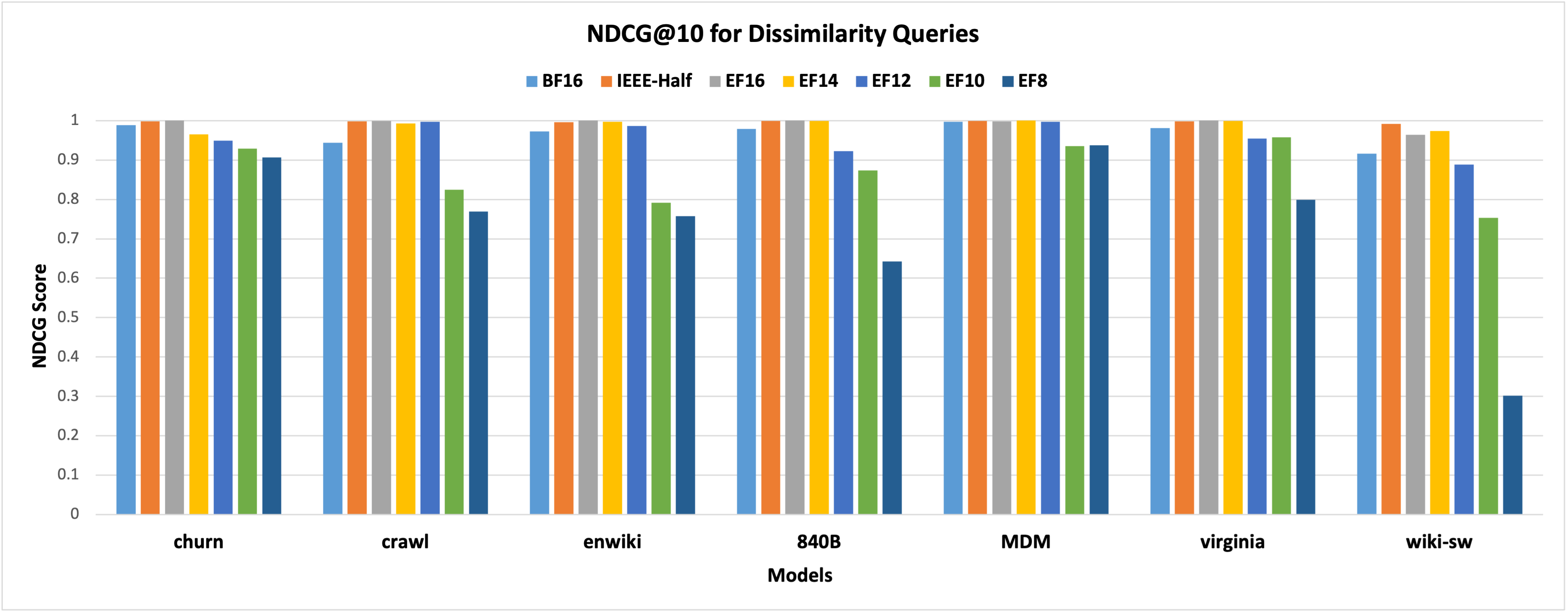}
  \caption{Evaluation of dissimilarity query accuracy using NDCG score across different floating point formats. Higher score (closer to 1.0) is better.}
  \label{fig:dissim}
\end{figure*}

Figure~\ref{fig:sim} presents NDCG\@10 results for similarity queries,
and Figure~\ref{fig:dissim} presents NDCG\@10 results for
dissimilarity queries. For both
similarity and dissimilarity queries, EF16 matches
or exceeds the quality of BF16 or FP16  (in particular,
among the three formats, BF16 provides the worst qaulity
results). Furthermore, EF14 and EF12 provide similar
quality results as EF16 in many instances. The two lower-precision EFn,
EF10 and EF8, consistently generate the least quality results.

In summary, results from the two sets of experiments
(Table~\ref{tab:errors}, and Figures~\ref{fig:sim} and
~\ref{fig:dissim}), conclusively demonstrate that: (1) Given a bit
budget, EFloat has higher accuracy than other formats,
(2) In
many scenarios, EFn with reduced bit budget (e.g.,
EF14 or EF12) provides results of quality comparable to higher
precision formats, e.g., BF16, and FP16. These results
validate the design of the EFloat format, and demonstrate that
EFloats can be used for compressing and computing using
vector embedding models.

%% file: related.tex
\section{Related Work}
\label{sec:related}

In this section, we overview relevant work in lossless compression techniques, deep learning model compression,
and floating point representations for deep learning.

Entropy encoding is a statistical method for lossless
compression~\cite{salomon2004data}.  Fixed-size items are replaced
with variable-size codes with the shorter codes assigned to the
frequently occurring items in the data.  Huffman coding, Arithmetic
coding and Range coding are commonly used entropy
methods. Variable-size codes, for example the Huffman codes, generally
have the {\em prefix-property} permitting their concatenation without
any separating markers in between.

Dictionary methods for lossless compression, most popularly the
Lempel-Ziv (LZ) algorithms, use a dictionary of
strings~\cite{salomon2004data}.  Strings in the data stream, when
found in the dictionary, are replaced with distance and length pairs
pointing to their dictionary location therefore achieving
compression. A {\em dynamic dictionary} is typically the most recent
set of input strings, e.g., the recent 32KB of input in the popular
Deflate method~\cite{zlib:library}.

Over the years, the size and complexity of deep learning models has
increased substantially. In particular, advent of new
transformer-based NLP models (e.g., BERT and friends, T5, Megatron-LM, Open AI GPT-2/3)
has highlighted the very high space and computational costs associated
with these models~\cite{rogers2020primer, gupta2020compression, may-blog, gordon-blog,
ganesh2020compressing, shoeybi2020megatronlm, bender:parrots, cacm:transformer-review}. Given potential uses of NLP models in
enterprise and consumer domains, a lot of attention is being devoted to
compressing such models. The primary goal of these compression
efforts is to reduce the size of a \emph{pre-trained} model to enable its
deployment in real world industrial applications
that demand low memory footprint, low response times, and smaller
computational and power budget during the \emph{inference} phase. Gupta and
Agarwal~\cite{gupta2020compression} have identified six different
types of compression techniques currently being used for the NLP
models: pruning, quantization, parameter sharing, knowledge
distillation, tensor decomposition, and Linear Transformer based
methods. The \emph{pruning} approach is the most obvious way to reduce
model size by sparsifying weight matrices. Pruning is related to
\emph{quantization} which aims to reduce the number of bits to
represent each weight. Quantization covers two broad approaches: the first
represents a full-precision (e.g., 32-bit) floating point weight value
using reduced or mixed precision representations, and the second
converts full-precision floating point values into integer values with
fewer bits (e.g., INT8, INT4, and INT1~\cite{int8-nvidia,
  wu2020integer, tensorflow-lite, jacob2017quantization} ). Another way to reduce
model size is \emph{parameter sharing} that uses fewer shared values
to represent similar weights. \emph{Knowledge distillation} aims to
build a student-trainer model where the student is trained to mimic a
pre-trained larger teacher model. The deeper teacher model is trained
first, and then the student model is trained via knowledge
transfer. \emph{Tensor factorization} covers a set of techniques that
can be used to approximate a larger matrix using a combination of
smaller matrices computed via tensor decomposition methods. The final
technique aims to develop transformer-based models that are
\emph{linear} in terms of input sequence size, rather than the current
quadratic complexity. In general, these techniques follow a
destructive approach that throws out the original large model and
can not recreate it from the compressed smaller model. 

Compression techniques have been explored to reduce data communication
costs during deep learning training. Floating-point quantization
approaches are often used to reduce the communication volume during
the deep learning training process~\cite{xu-compression}. Gajjala et
al~\cite{huffman-quant} use Huffman encoding based techniques to
encode quantized gradients for optimizing communication volume in
distributed deep learning training. Recently
announced Nvidia nvcomp~\cite{nvcomp} uses LZ4 and run-length encoding (RLE)
based approaches to compress data being communicated between GPUs during training
of deep learning models.  While the approaches used
in nvcomp work well for string and integer datatypes, these techniques
do not support floating point values very well. Also, LZ4 and RLE
approaches work well only for repeated values or sequences.
The EFloat approach is compatible and orthogonal to these existing compression
techniques employed in NLP models.

In conjunction with the model compression work, there has been
significant activity in devising reduced-precision floating point formats tuned
for broader machine learning and HPC applications~\cite{model-zoo,
  abdelfattah2020survey}. Unlike the inference-focused model
compression work, reduced-precision floating point representations are
designed to work for both model training and inference phases. The
most common reduced-precision floating point representation uses 16
bits. Current 16-bit implementations include IEEE 754 half-precision
(FP16), with 1 sign bit, 5 exponent bits, and 10 fraction bits; Brain
Floating Point (BFLOAT16)~\cite{wang:bfloat16, kalamkar2019study},  with 1 sign bit, 8 exponent bits, and 7
fraction bits; and Deep Learning Float (DLFloat)~\cite{DLFloat}, with 1 sign bit, 6
exponent bits, and 7 fraction bits. TensorFloat-32 (TF32) from Nvidia
is a 19-bit format that combines 8 exponent bits from BFLOAT16 and 10
exponent bits from IEEE FP16. Hybrid Block Floating Point (HBFP)~\cite{hbfp-neurips}, Intel Nervana's
Flexpoint~\cite{flexpoint}, and Microsoft
MSFP~\cite{rouhani2020pushing} formats
combine the advantages of fixed point and floating point
representations by splitting up the mantissa and the exponent part
which is shared across multiple numeric values. Recent research
proposals have described training of key deep learning models using
even reduced precision floating point values (8- and 4-bit
representations)~\cite{NEURIPS2020-4bit, NEURIPS2018-8bit, cambier2020shifted, mellempudi2019mixed}. 
Recently proposed AdaptiveFloat~\cite{adaptiveFloat-dac} and AdaptivePosit~\cite{langroudi-posit} are inference-targeted floating-point formats
that maximize their dynamic range \emph{per network layer} by dynamically shifting its exponent range via modifications
to the exponent bias and by optimally clipping (quantizing) its representable datapoints. Our proposed EFloat design
practically achieves the same result without altering the exponent range and quantizing full-precision values, and it does not need 
to change per neural network layer.


%% file: concl.tex
\section{Conclusion}
\label{sec:concl}
We introduced EFloat, a novel entropy-coded variable length floating
point format for deep learning applications. This format can be used
for compressing a trained deep learning model, as well as for enabling
more accurate model representations using reduced-precision floating point
formats. While our intended use cases were
initially for the database embedding (db2Vec) workloads, we
demonstrate that the proposed format works effectively for other
vector embedding models, and can be used for a much broader class of
NLP models including transformer-based models. Broadly, EFloat may be
used in deep learning applications where
tradeoffs need to be made between range, precision, memory capacity
and bandwidth savings. As a future work, we plan to explore the Benford distribution
pattern~(\citet{benford-law,newcomb-law}) exhibited by significands of vector embedding models (Section~\ref{sec:impl})
and investigate its application in rounding EFloat values.
A follow-up study on 8-bit floats and integers is being considered as well.

%% file: ack.tex
\section{Acknowledgments}
We thank Jose Neves, Jeff Burns, and Kathryn O'Brien for their insightful comments on the earlier versions of this paper.